\documentclass[sigconf, nonacm]{acmart}

\usepackage{verbatim}
\usepackage{placeins}
\usepackage{multirow}
\usepackage{textcomp}
\usepackage{amsmath}
\usepackage{amssymb}
\usepackage{enumitem}
\usepackage{setspace}
\usepackage{subcaption}
\usepackage{natbib}
\usepackage{pgf}

\usepackage{algorithm}
\usepackage[noend]{algpseudocode}
\makeatletter
\renewcommand{\ALG@beginalgorithmic}{\small}
\makeatother
\algnewcommand\Input{\textbf{Input:\space}}
\algnewcommand\Output{\textbf{Ouput:\space}}

\usepackage{multicol}
\usepackage{tikz}
\usepackage{pgf}
\usepackage[capitalise]{cleveref}
\labelcrefformat{equation}{#2\textup{#1}#3}
\creflabelformat{equation}{#2#1#3}
\crefrangelabelformat{equation}{#3#1#4 to #5#2#6}

\newcommand{\COMMENT}[1]{}

\usepackage{glossaries}

\newacronym{gps}
{GPS}
{Global Positioning System}

\newacronym{osm}
{OSM}
{OpenStreetMap}

\newacronym{rne}
{RNE}
{Road Network Embedding}

\newacronym{dfs}
{DFS}
{Depth-First Search}

\newacronym{elu}
{ELU}
{Exponential Linear Unit}

\newacronym{relu}
{ReLU}
{Rectified Linear Unit}

\newacronym{bfs}
{BFS}
{Breadth-First Search}

\newacronym{gcn}
{GCN}
{Graph Convolutional Network}

\newacronym{rfn}
{RFN}
{Relational Fusion Network}

\newacronym{mlp}
{MLP}
{Multi-Layer Perceptron}

\newacronym{tsne}
{t-SNE}
{t-distributed Stochastic Neighbor Embedding}

\newacronym{mape}
{MAPE}
{Mean Absolute Percentage Error}

\newacronym{mae}
{MAE}
{Mean Absolute Error}

\newacronym{mse}
{MSE}
{Mean Squared Error}

\newacronym{mpe}
{MPE}
{Mean Percentage Error}

\usepackage[disable]{todonotes}
\presetkeys{todonotes}{inline}{}


\setcopyright{none}

\acmDOI{10.475/123_4}

\acmISBN{123-4567-24-567/08/06}
\acmSubmissionID{77}

\acmConference[SIGSPATIAL'19]{ACM SIGSPATIAL}{November 5--8 2019}
{Chicago, IL, USA}
\acmYear{2019}
\copyrightyear{2019}

\begin{document}
\begin{abstract}
  Machine learning techniques for road networks hold the potential to facilitate many important transportation applications.
  \glspl{gcn} are neural networks that are capable of leveraging the structure of a road network by utilizing information of, e.g., adjacent road segments.
  While state-of-the-art \glspl{gcn} target node classification tasks in social, citation, and biological networks, machine learning tasks in road networks differ substantially from such tasks. In road networks, prediction tasks concern edges representing road segments, and many tasks involve regression.
  In addition, road networks differ substantially from the networks assumed in the \gls{gcn} literature in terms of the attribute information available and the network characteristics. Many implicit assumptions of \glspl{gcn} do therefore not apply.
 
  We introduce the notion of \emph{\glsfirst{rfn}}, a novel type of \gls{gcn} designed specifically for machine learning on road networks.
  In particular, we propose methods that outperform state-of-the-art \glspl{gcn} on both a road segment regression task and a road segment classification task by $32\text{--}40\%$ and $21\text{--}24\%$, respectively.
  In addition, we provide experimental evidence of the short-comings of state-of-the-art \glspl{gcn} in the context of road networks: unlike our method, they cannot effectively leverage the road network structure for road segment classification and fail to outperform a regular multi-layer perceptron.
\end{abstract}

\title{Graph Convolutional Networks for Road Networks}
\settopmatter{authorsperrow=3}
\author{Tobias Skovgaard Jepsen}
\affiliation{%
  \institution{Department of Computer Science}
  \institution{Aalborg University}
}
\email{tsj@cs.aau.dk}
\author{Christian S. Jensen}
\affiliation{%
  \institution{Department of Computer Science}
  \institution{Aalborg University}
}
\email{csj@cs.aau.dk}

\author{Thomas Dyhre Nielsen}
\affiliation{%
  \institution{Department of Computer Science}
  \institution{Aalborg University}
}
\email{tdn@cs.aau.dk}

\maketitle

\section{Introduction}\label{sec:introduction}
Machine learning on road networks can facilitate important transportation applications such as traffic forecasting~\citep{spatio-temporal-gcn}, speed limit annotation~\citep{workshop}, and travel time estimation.
However, machine learning on road networks is difficult due to the low number of attributes, often with missing values, that typically are available~\citep{workshop}.
This lack of attribute information can be alleviated by exploiting the network structure into the learning process~\citep{workshop}.
To this end, we propose the \emph{\glsfirst{rfn}}, a type of \glsfirst{gcn} designed for machine learning on road networks.

\glspl{gcn} are a type of neural network that operates directly on graph representations of networks.
\glspl{gcn} can in theory leverage the road network structure since they compute a representation of a road segment by aggregating over its neighborhood, e.g., by using the mean representations of its adjacent road segments.
However, state-of-the-art \glspl{gcn} are designed to target node classification tasks in social, citation, and biological networks.
Although \glspl{gcn} have been highly successful at such tasks, machine learning tasks in road networks differ substantially: they are typically prediction tasks on edges and many important tasks are regression tasks.

Road networks differ substantially from networks studied in the \gls{gcn} literature in terms of the attribute information available and network characteristics. Many implicit assumptions in \gls{gcn} proposals do not hold. 
First, road networks are \emph{edge-relational} and contain not only node and edge attributes, but also \emph{between-edge attributes} that characterize the relationship between road segments, i.e., the edges in a road network.
For instance, the angle between two road segments is informative for travel time estimation since it influences the time it takes to move from one segment to the other.

\begin{figure}[t]
  \centering
    \includegraphics[trim={0cm 6cm 1.5cm 8cm},clip, width=0.95\columnwidth]{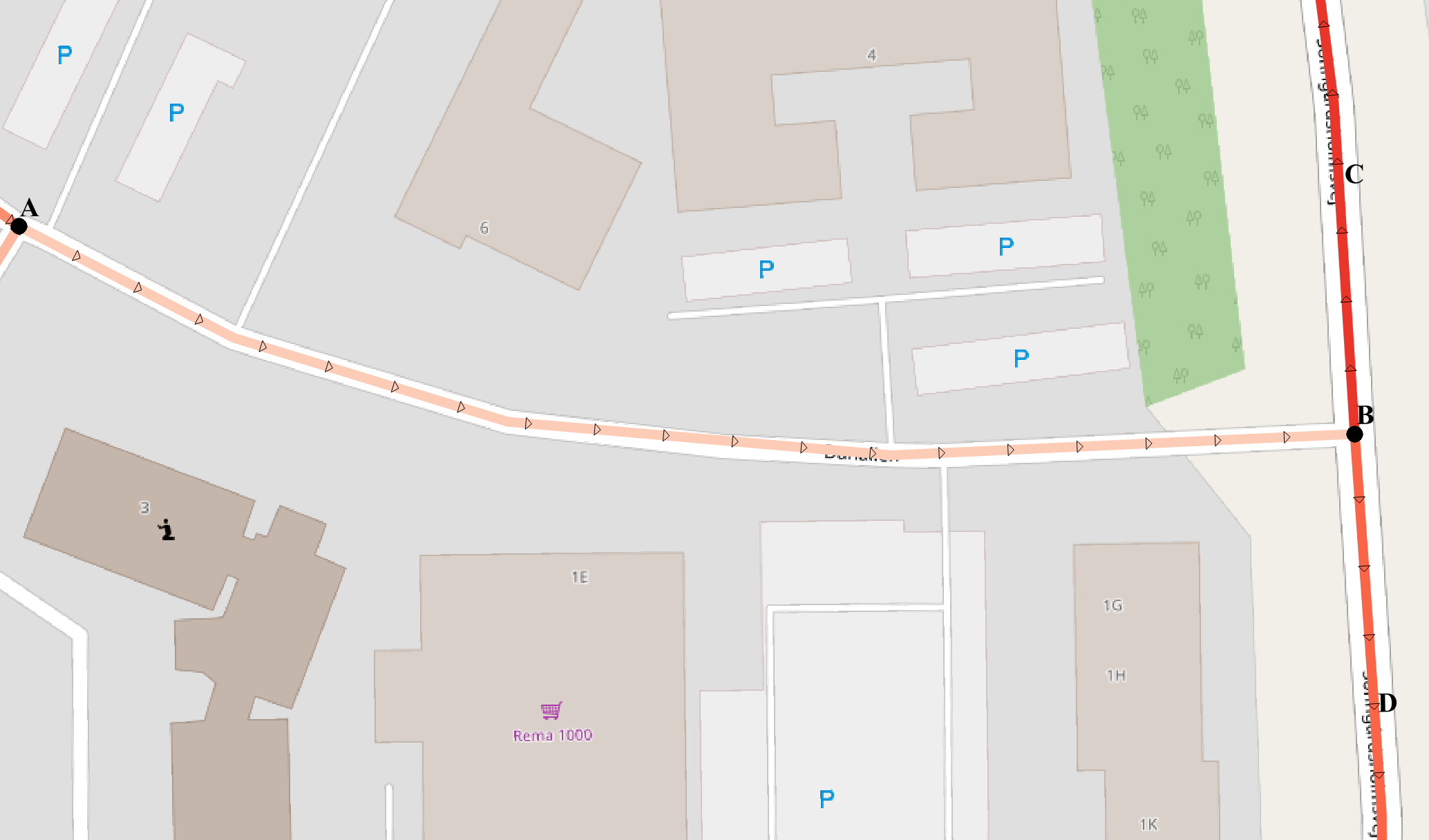}
  \caption{Two three-way intersections in Denmark. We illustrate the observed driving speeds for one driving direction per segment and color them accordingly: the darker, the faster the speed. Black dots mark intersections and triangles indicate driving directions.\label{fig:compelling-example}}
\end{figure}

Second, compared to social, citation, and biological networks, road networks are low-density: road segments have few adjacent road segments.
For instance, the Danish road network has a mean node degree of $2.2$~\citep{workshop} compared to the mean node degrees $9.15$ and $492$ of a citation and a social networks, respectively~\citep{graphsage}.
The small neighborhoods in road networks make neighborhood aggregation in \glspl{gcn} sensitive to aberrant neighbors.

Third, \glspl{gcn} implicitly assume that the underlying network exhibits homophily, e.g., that adjacent road segments tend to be similar, and that changes in network characteristics, e.g., driving speeds, occur gradually.
Although road networks exhibit homophily, the homophily is volatile in the sense that regions can be highly homophilic, but have sharp boundaries characterized by abrupt changes in, e.g., driving speeds.
In the most extreme case, a region may consist of a single road segment, in which case there is no homophily.
As an example, the three-way intersection to the right in \cref{fig:compelling-example} exhibits volatile homophily.
The two vertical road segments to the right in the figure and the road segments to the left in the figure form two regions that are internally homophilic: the road segments in the regions have similar driving speeds to other road segments in the same region.
These regions are adjacent in the network, but, as indicated by the figure, a driver moving from one region to the other experiences an abrupt change in driving speed.

\paragraph{Contributions.} We introduce the notion of \emph{\glsfirst{rfn}}, a type of \gls{gcn} designed specifically to address the shortcomings of state-of-the-art \glspl{gcn} in the road network setting.
At the core of a \gls{rfn} is the relational fusion operator, a novel graph convolutional operator that aggregates over representations of relations instead of over representations of neighbors.
Furthermore, it is capable of disregarding relations with aberrant neighbors during the aggregation through an attention mechanism.
In combination, these features allow \glspl{rfn} to extract only the useful information from only useful relations.

We treat the learning of relational representations during relational fusion as a multi-view learning problem where the objective is to learn a relational representation of a relation $(u, v)$ indicating a relationship between, e.g., two road segments $u$ and $v$.
These representations are learned by fusing the representations, e.g., attributes, of road segment $u$ and $v$ and the attributes of the relation $(u, v)$ that describe the nature of the relationship between $u$ and $v$.

As a key feature, \glspl{rfn} consider both the relationships between intersections (nodes) and between road segments (edges) jointly, i.e., they adopt both a \emph{node-relational} and an \emph{edge-relational} view.
These two views are interdependent and by learning according to both perspectives, an \gls{rfn} can utilize node attributes, edge attributes, and \emph{between-edge attributes} jointly during the learning process.
In contrast, state-of-the-art \glspl{gcn}~\citep{bruna2014, chebnet, gcn, graphsage, gat} may learn only from one of these perspectives, using either node or edge attributes.
In addition, the edge-relational view inherently focuses on making predictions (or learning representations) of edges representing road segments which we argue is the typical object of interest in machine learning tasks in road networks.

We evaluate different variants of the proposed \glspl{rfn} on two road segment prediction tasks: driving speed estimation and speed limit classification in the Danish municipality of Aalborg.
We compare these \glspl{rfn} against two state-of-the-art \glspl{gcn}, a regular multi-layer perceptron, and a naive baseline.
We find that all \gls{rfn} variants outperform all baselines, and the best variants outperform state-of-the-art \glspl{gcn} by $32\text{--}40\%$ and $21\text{--}24\%$ on driving speed estimation and speed limit classification, respectively.
Interestingly, we also find that the \gls{gcn} baselines are unable to outperform the multi-layer perceptron on the speed limit classification task, unlike our method.
This indicates that the \glspl{gcn} cannot leverage neighborhood information from adjacent road segments on this task despite such information being present as indicated by the superior results of our method.

The remainder of the paper is structured as follows.
In \cref{sec:related-work}, we review related work.
In \cref{sec:preliminaries}, we give the necessary background on graph modeling of road networks and \glspl{gcn}.
In \cref{sec:method}, we describe \glspl{rfn} in detail.
In \cref{sec:experiments}, we describe our experiments and present our results.
Finally, we conclude in \cref{sec:conclusion}.

\section{Related Work}\label{sec:related-work}
In earlier work, \citet{workshop} investigate the use of network embedding methods for machine learning on road networks. Network embedding methods are closely related to \glspl{gcn}, and both rely on the core assumption of homophily in the network.
They find that homophily is expressed differently in road networks---which we refer to as volatile homophily---but study this phenomenon in the context of transductive network embedding methods, whereas we focus on inductive \glspl{gcn}.
Unlike us, they do not propose a method to address the challenges of volatile homophily in road networks.

\citet{spatio-temporal-gcn} extend \glspl{gcn} to the road network setting but focus on the temporal aspects of road networks rather than the spatial aspects which are the focus of our work. As such, the method we propose is orthogonal and complimentary to theirs and we expect that our proposed \glspl{rfn} can be used as a replacement for the spatial convolutions of their method.
Furthermore, their work focuses on the specific task of traffic forecasting whereas we propose a general-purpose method for performing machine learning on road networks.
Finally, their method belongs to the class of spectral \glspl{gcn}~\citep{bruna2014, chebnet, gcn}, whereas our method belongs to the class of spatial \glspl{gcn}~\citep{graphsage, gat}.

Neither state-of-the-art spectral \glspl{gcn}~\citep{bruna2014, chebnet, gcn} nor spatial \glspl{gcn}~\citep{graphsage, gat} support relational attributes between intersections and road segments, such as edge attributes and between-edge attributes.
Furthermore, spectral \glspl{gcn} are defined only for undirected graphs, whereas road networks are typically directed, e.g., they contain one-way streets.
In contrast, our method can incorporate edge attributes and between-edge attributes, in addition to node attributes, and it learns on a directed graph representation of a road network.

Unlike spectral \glspl{gcn}, spatial \glspl{gcn} (such as the \glspl{rfn} we propose) are inductive and are therefore capable of transferring knowledge from one region in a road network to another region or even from one road network to another.
The inductive nature of spatial \glspl{gcn} also makes training them less resource-intensive since the global gradient can be approximated from small (mini-)batches of a few, e.g., road segments, rather than performing convolution on the whole graph.
Our proposed \glspl{rfn} are compatible with existing mini-batch training algorithms for spatial \glspl{gcn}~\citep{graphsage,fastgcn}.

\section{Preliminaries}\label{sec:preliminaries}
We now cover the necessary background in modeling road networks as graphs and \glspl{gcn}.

\subsection{Modeling Road Networks}\label{sec:road-network-modelling}
We model a road network as an attributed directed graph $G = (V, E, A^V, A^E, A^B)$ where $V$ is the set of nodes and $E$ is the set of edges.
Each node $v \in V$ represents an intersection, and each edge $(u, v) \in E$ represents a road segment that enables traversal from $u$ to $v$.
Next, $A^V$ and $A^E$ maps intersections and road segments, respectively, to their attributes. In addition, $A^B$ maps a pair of road segments $(u, v), (v, w) \in E$ to their between-segment attributes such as the angle between $(u, v)$ and $(v, w)$ based on their spatial representation.
An example of a graph representation of the three-way intersection to the right in figure \cref{fig:compelling-example} is shown in \cref{fig:graph-rep-primal}.

\begin{figure}
  \centering
  \begin{subfigure}{0.495\columnwidth}
    \centering
    \usetikzlibrary{shapes,arrows,positioning,fit,backgrounds}
\begin{tikzpicture}[align=center]
  \node [draw, circle] (A) {A};
  \node [draw, circle, right=1cm of A] (B) {B};
  \node [draw, circle, above=1cm of B] (C) {C};
  \node [draw, circle, below=1cm of B] (D) {D};
  
  \draw[-latex] (A) to [out=15, in=165] (B);
  \draw[-latex] (B) to [out=195, in=345] (A);
  
  \draw[-latex] (B) to [out=105, in=255] (C);
  \draw[-latex] (C) to [out=285, in=75] (B);
  
  \draw[-latex] (B) to [out=285, in=75] (D);
  \draw[-latex] (D) to [out=105, in=255] (B);
\end{tikzpicture}
\vspace{-0.5cm}
    \vspace*{0.5cm}
    \caption{Primal Graph.\label{fig:graph-rep-primal}}
  \end{subfigure}
  \begin{subfigure}{0.495\columnwidth}
    \centering
    \vspace*{0.6825cm}
    \usetikzlibrary{shapes,arrows,positioning,fit,backgrounds}
\usetikzlibrary{positioning}
\begin{tikzpicture}[align=center]
  \node [draw, circle] (AB) {AB};
  \node [draw, circle, below=1cm of AB] (BA) {BA};
  
  \node [draw, circle, right=1cm of AB] (BC) {BC};
  \node [draw, circle, right=1cm of BC] (CB) {CB};

  \node [draw, circle, right=1cm of BA] (DB) {DB};
  \node [draw, circle, right=1cm of DB] (BD) {BD};
  
  \draw[-latex] (AB) to [out=285, in=75] (BA);
  \draw[-latex] (BA) to [out=105, in=255] (AB);
  \draw[-latex] (AB) -- (BD);
  \draw[-latex] (AB) -- (BC);

  \draw[-latex] (BC) to [out=15, in=165] (CB);
  \draw[-latex] (CB) to [out=195, in=345] (BC);

  \draw[-latex] (DB) to [out=15, in=165] (BD);
  \draw[-latex] (BD) to [out=195, in=345] (DB);

  \draw[-latex] (CB) -- (BA);
  \draw[-latex] (CB) -- (BD);

  \draw[-latex] (DB) -- (BA);
  \draw[-latex] (DB) -- (BC);

  
  
\end{tikzpicture}
\vspace{-0.5cm}
    \vspace*{0.6825cm}
    \caption{Dual Graph.\label{fig:graph-rep-dual}}
  \end{subfigure}
  \caption{The (a) primal and (b) dual graph representations of the three-way intersection to the right in \cref{fig:compelling-example}.\label{fig:graph-rep}}
  \vspace*{-0.5cm}
\end{figure}

Two intersections $u$ and $v$ in $V$ are adjacent if there exists a road segment $(u, v) \in E$ or $(v, u) \in E $. 
Similarly, two road segments $(u_1, v_1)$ and $(u_2, v_2)$ in $E$ are adjacent if $v_1 = u_2$ or $v_2 = u_1$. 
The function $N\colon V \cup E \xrightarrow{} 2^V \cup 2^E$ returns the neighborhood, i.e., the set of all adjacent intersections or road segments, of a road network element $g \in V \cup E$.
The dual graph representation of $G$ is then $G^D=(E, B)$ where $B = \big\{ \big((u, v), (v, w) \big) \mid (u, v), (v, w) \in E \big\}$ is the set of \emph{between-edges}.
Thus, $E$ and $B$ are the node and edge sets, respectively, in the dual graph. 
For disambiguation, we refer to $G$ as the primal graph representation.

\subsection{Graph Convolutional Networks}\label{sec:gcn}
A \gls{gcn} is a neural network that operates on graphs and consists of one or more graph convolutional layers.
A graph convolutional network takes as input a graph $G = (V, E)$ and a numeric node feature matrix $\mathbf{X}^V \in \mathbb{R}^{|V| \times d_{\mathit{in}}}$, where each row corresponds to a $d_{\mathit{in}}$-dimensional vector representation of a node. 
Given these inputs, a \gls{gcn} computes an output at a layer $k$ s.t.\
\begin{equation}\label{eq:gcn-propagation}
  \mathbf{H}^{(V, k)}_v = \sigma(\textsc{Aggregate}^k(v) \mathbf{W})\text{,}
\end{equation}
where $\sigma$ is an activation function, and $\textsc{Aggregate}\colon 2^V \rightarrow \mathbb{R}^{d_{\mathit{in}}}$ is an aggregate function.
Similarly to $\mathbf{X}^V$, each row in $\mathbf{H}^{(V, k)}$ is a vector representation of a node.
Note that in some cases $\mathbf{X}^V$ is linearly transformed using matrix multiplication with a weight matrix $\mathbf{W}$ before aggregation~\citep{gat} while in other cases, weight multiplication is done after aggregation~\citep{gcn,graphsage}, as in \cref{eq:gcn-propagation}.

The \textsc{Aggregate} function in \cref{eq:gcn-propagation} derives a new representation of a node $v$ by aggregating over the representations of its neighbors.
While the aggregate function is what distinguishes \glspl{gcn} from each other, many can be expressed as a weighted sum~\citep{gcn,graphsage,gat}:
\begin{equation}\label{eq:aggregateWeightedSum}
  \textsc{Aggregate}^k(v) = \sum_{n \in N(v)} a_{(v,n)} \mathbf{H}^{(V, k-1)}_n\text{,}
\end{equation}
where $\mathbf{H}^{(V, 0)} = \mathbf{X}^V$, and $a_{(v,n)}$ is the aggregation weight for neighbor $n$ of node $v$.
As a concrete example, the mean aggregator of GraphSAGE~\citep{graphsage} uses $a_{(v, n)} = |N(v)|^{-1}$ in \cref{eq:aggregateWeightedSum}.

\section{Relational Fusion Networks}\label{sec:method}
\emph{\glsfirst{rfn}} aim to address the shortcomings of state-of-the-art \glspl{gcn} in the context of machine learning on road networks. The basic premise is to learn representations based on two distinct, but interdependent views: the node-relational and edge-relational views.

\subsection{Node-Relational and Edge-Relational Views}
In the node-relational view, we seek to learn representations of nodes, i.e., intersections, based on their node attributes and the relationships between nodes indicated by the edges $E$ in the primal graph representation of a road network $G^P = (V, E)$ and described by their edge attributes.
Similarly, we seek to learn representations of edges, i.e., road segments, in the edge-relational view, based on their edge attributes and the relationships between edges indicated by the between-edges $B$ in the dual graph representation of a road network $G^D = (E, B)$.
The relationship between two adjacent roads $(u, v)$ and $(v, w)$ is described by the attributes of the between-edge connecting them in the dual graph, including the angle between them, but also the attributes of the node $v$ that connects them.

The node- and edge-relational views are complementary.
The representation of a node in the node-relational view is dependent on the representation of the edges to its neighbors.
Similarly, the representation of an edge in the edge-relational view is dependent on the representation of the nodes that it shares with their neighboring edges.
Finally, the representation of an edge is also dependent on the representation of the between-edge connecting them in the dual graph.
\glspl{rfn} can exploit these two complementary views to leverage node, edge, and between-edge attributes simultaneously.


\subsection{Method Overview}
\begin{figure}
  \centering
  \begin{subfigure}{\columnwidth}
    \centering
    \usetikzlibrary{shapes,arrows,positioning,fit,backgrounds,calc}
\begin{tikzpicture}[
    align=center,
    layer/.style={shape=rectangle, text width=2.25cm, align=center, draw=black},
    arrow/.style={-latex, thick}
  ]
   \def\layerdistance{1.1cm}
   \def\layeroffset{1.75}
   \def\inoutoffset{0.25cm}

    \node[] (Xn) at (0.75,-\inoutoffset) {$\mathbf{X}^V$};
    \node[] (Xe) at (5,-\inoutoffset) {$\mathbf{X}^E$};
    \node[] (Xbe) at (6,-\inoutoffset) {$\mathbf{X}^{B}$};
   
    \node [draw, fill=blue!50!white!10, rounded corners, minimum width=7.5cm, minimum height=1.5cm, outer sep=0, inner sep=0] at (2.9,1.09) {\textit{Layer $1$}};
  
    \node[layer, fill=yellow] (GCN1n) at (0.75,1*\layerdistance) {Node-Relational Fusion};
    \draw[arrow] (Xn) -- (GCN1n);
    \draw[arrow] (Xe) -- (GCN1n);
    
    \node[layer, fill=yellow] (GCN1e) at (5,1*\layerdistance) {Edge-Relational Fusion};
    \draw[arrow] (Xn) -- (GCN1e);
    \draw[arrow] (Xe) -- (GCN1e);
    \draw[arrow] (Xbe) -- (GCN1e);

    \node [draw, fill=blue!50!white!10, rounded corners, minimum width=7.5cm, minimum height=1.5cm, outer sep=0, inner sep=0] at (2.9,3.09) {\textit{Layer $2$}};

    \node[layer, fill=yellow] (GCN2n) at (0.75,2cm + 1*\layerdistance) {Node-Relational Fusion};
    \draw[arrow] (GCN1n) -- (GCN2n);
    \draw[arrow] (GCN1e) -- (GCN2n);
    
    \node[layer, fill=yellow] (GCN2e) at (5,2cm + 1*\layerdistance) {Edge-Relational Fusion};
    \draw[arrow] (GCN1e) -- (GCN2e);
    \draw[arrow] (GCN1n) -- (GCN2e);
   
    \node [draw, fill=blue!50!white!10, rounded corners, minimum width=7.5cm, minimum height=1.5cm, outer sep=0, inner sep=0] (LK) at (2.9,5.09 + 0.5) {\textit{Layer $K$}};

    \node[layer, fill=yellow] (GCNkn) at (0.75,4.5cm + 1*\layerdistance) {Node-Relational Fusion};
    \node[draw=none] (A3) at (2.3, 4.801) {};
    \draw[arrow] (A3.south) -- (GCNkn);
    \draw[thick] (GCN2e) edge[] ($(GCN2e)!0.575!(A3.south)$) edge [dotted] ($(GCN2e)!1.0!(A3.south)$);
    
    \node[draw=none, above=1.2cm of GCN2n] (A1) {};
    \draw[arrow] (A1.south) -- (GCNkn);
    \draw[thick] (GCN2n) edge[] ($(GCN2n)!0.65!(A1.south)$) edge [dotted] ($(GCN2n)!1.0!(A1.south)$);
    
    \node[layer, fill=yellow] (GCNke) at (5,4.5cm + 1*\layerdistance) {Edge-Relational Fusion};
    \node[draw=none, above=1.2cm of GCN2e] (A2) {};
    \draw[arrow] (A2.south) -- (GCNke);
    \draw[thick] (GCN2e) edge[] ($(GCN2e)!0.65!(A2.south)$) edge [dotted] ($(GCN2e)!1.0!(A2.south)$);

    \node[draw=none] (A4) at (3.5, 4.83) {};
    \draw[arrow] (A4.south) -- (GCNke);
    \draw[thick] (GCN2n) edge[] ($(GCN2n)!0.575!(A4.south)$) edge [dotted] ($(GCN2n)!1.0!(A4.south)$);
    \node[above=\layerdistance-\inoutoffset of GCNkn] (Zn) {$\mathbf{H}^{(V, K)}$};
    \draw[arrow] (GCNkn) -- (Zn);
    
    \node[above=\layerdistance-\inoutoffset of GCNke] (Ze) {$\mathbf{H}^{(E, K)}$};
    \draw[arrow] (GCNke) -- (Ze);
    
    \node[right=0cm of Ze] (Zb) {$\mathbf{H}^{(B, K)}$};
    \draw[arrow] (GCNke) -- (Zb);
    
\end{tikzpicture}
\vspace{-0.5cm}
    \vspace*{0.5cm}
    \caption{Relational Fusion Network\label{fig:methodOverviewA}}
  \end{subfigure}
  \begin{subfigure}{\columnwidth}
    \centering
    \usetikzlibrary{shapes,arrows,positioning,fit,backgrounds}
\begin{tikzpicture}[
    align=center,
    layer/.style={shape=rectangle, text width=1.8cm, align=center, draw=black, minimum height=0.875cm},
    arrow/.style={-latex, thick}
  ]
    \node [draw, fill=blue!50!white!10, rounded corners, minimum width=8.5cm, minimum height=1.6cm, outer sep=0, inner sep=0] at (2.25,1.22) {};
    
    \node [draw, fill=yellow, rounded corners, minimum width=3cm, minimum height=1.1cm, outer sep=0, inner sep=0] at (-0.4,1.215) {};
    \node [opacity=0, text opacity=1, outer sep=0, inner sep=0, text width=2.5cm, align=left] at (-0.575,1.215) {\tiny{Node-\vspace{-0.1cm}\break Relational\vspace{-0.1cm}\break Fusion}};
  
    \node [draw, fill=yellow, rounded corners, minimum width=5.25cm, minimum height=1.1cm, outer sep=0, inner sep=0] at (3.775,1.215) {};
    \node [opacity=0, text opacity=1, outer sep=0, inner sep=0, text width=1cm, align=left] at (6.1,1.215) {\tiny{Edge-\vspace{-0.1cm}\break Relational\vspace{-0.1cm}\break Fusion}};

   \def\layerdistance{0.5cm}
   \def\inoutoffset{0.25cm}
    \node[] (Xn) at (0,-\inoutoffset) {$\mathbf{H}^{(V, k-1)}$};
    \node[] (Xe) at (2.25,-\inoutoffset) {$\mathbf{H}^{(E, k-1)}$};
    \node[] (Xbe) at (4.5,-\inoutoffset) {$\mathbf{H}^{(B, k-1)}$};

    \node[layer, above=\layerdistance + \inoutoffset of Xn, fill=cyan!255!white!25] (GCN1n) {\footnotesize{Relational Fusion\\(Primal Graph)}};
    \draw[arrow] (Xn) -- (GCN1n);
    \draw[arrow] (Xe) -- (GCN1n);
    
    \node[layer, above=\layerdistance + \inoutoffset of Xe, fill=cyan!255!white!25] (GCN1e) {\footnotesize{Relational Fusion\\(Dual Graph)}};
    \draw[arrow] (Xn) -- (GCN1e);
    \draw[arrow] (Xe) -- (GCN1e);
    \draw[arrow] (Xbe) -- (GCN1e);

    \node[layer, above=\layerdistance + \inoutoffset of Xbe, fill=gray!255!white!20] (FF1be) {\footnotesize{Feed\\Forward}};
    \draw[arrow] (Xbe) -- (FF1be);
    
    \node[above=\layerdistance + \inoutoffset of GCN1n] (Zn) {$\mathbf{H}^{(V, k)}$};
    \draw[arrow] (GCN1n) -- (Zn);
    
    \node[above=\layerdistance + \inoutoffset of GCN1e] (Ze) {$\mathbf{H}^{(E, k)}$};
    \draw[arrow] (GCN1e) -- (Ze);
    
    \node[above=\layerdistance + \inoutoffset of FF1be] (Zbe) {$\mathbf{H}^{(B, k)}$};
    \draw[arrow] (FF1be) -- (Zbe);
\end{tikzpicture}
\vspace{-0.5cm}
    \caption{Relational Fusion Layer\label{fig:methodOverviewB}}
  \end{subfigure}
  \caption{Overview of our method showing (a) a $K$-layered relational fusion network and (b) a relational fusion layer.\label{fig:methodOverview}}
  \vspace*{-0.1cm}
\end{figure}
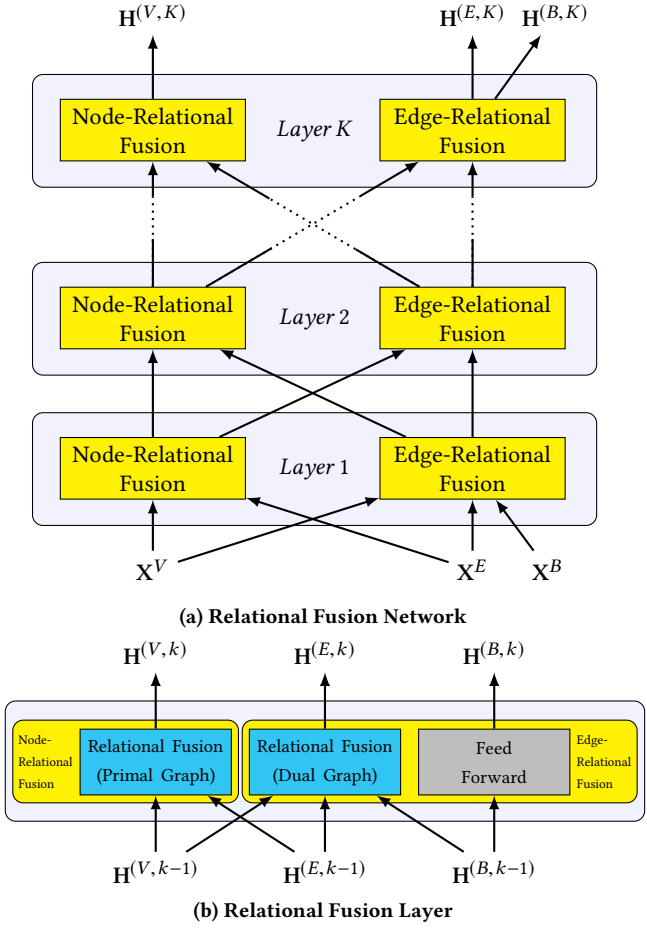

\cref{fig:methodOverviewA} gives an overview of our method.
As shown, an \gls{rfn} consists of $K$ relational fusion layers, where $K \geq 1$.
It takes as input feature matrices $\mathbf{X}^V \in \mathbb{R}^{|V \times d^V}$, $\mathbf{X}^E \in \mathbb{R}^{|E| \times d^E}$, and $\mathbf{X}^B \in \mathbb{R}^{|B| \times d^B}$ that numerically encode the node, edge, and between-edge attributes, respectively.
These inputs are then propagated through each layer.
Each of these relational fusion layers, performs \emph{node-relational fusion} and \emph{edge-relational fusion} to learn representations from the node- and edge-relational views, respectively. 

Node-relational fusion is carried out by performing relational fusion on the primal graph. Relational fusion is a novel graph convolutional operator that we describe in detail in \cref{sec:relationalFusion}.
In brief, relational fusion performs a graph convolution where the neighborhood aggregate is replaced by a relational aggregate.
When performing a relational aggregate for, e.g., a node $v$ in the primal graph, aggregation is performed over representations of the relations $(v, n)$ that $u$ participates in as opposed to over the representations of its neighbors.
These relational representations are a fusion of the node representations of $v$ and $n$, but also the edge representation of $(v, n)$.

Edge-relational fusion is performed similarly to node-relational fusion but is instead applied on the dual graph representation.
Recall, that in the edge-relational representation, the relations between two edge $(u, v)$ and $(v, w)$ is in part described by their between-edge attributes, but also by the node attributes of $v$ that describes the characteristics of the intersection between them.
Therefore, as illustrated in \cref{fig:methodOverviewB}, relational fusion on the dual graph requires both node and between-edge information to compute relational aggregates.
Finally, we apply regular feed-forward neural network propagation on the between-edge representations at each layer to learn increasingly abstract between-edge representations.
Note that we capture the interdependence between the node- and edge-relational views by using the node and edge representations from the previous layer $k-1$ as input to node-relational and edge-relational fusion in the next layer $k$ as illustrated by \cref{fig:methodOverview}.

\cref{fig:methodOverviewB} gives a more detailed view of a relational fusion layer.
Each layer $k$ takes as input the learned node, edge, and between-edge representations from layer $k-1$, denoted by $\mathbf{H}^{(V, k-1)}$, $\mathbf{H}^{(E, k-1)}$, and $\mathbf{H}^{(B, k-1)}$, respectively.
Then node-relational and edge-relational fusion are performed to output new node, edge, and between-edge representations $\mathbf{H}^{(V, k)}$, $\mathbf{H}^{(E, k)}$, and $\mathbf{H}^{(B, k)}$.

\subsection{Relational Fusion}\label{sec:relationalFusion}
We present the pseudocode for the relational fusion operator at the $k$th layer in \cref{alg:relationalFusion}.
The operator takes as input a graph $G' = (V', E')$, that is either the primal or dual graph representation of a road network, along with appropriate feature matrices $\mathbf{H}^{(V', k-1)}$ and $\mathbf{H}^{(E', k-1)}$ to describe nodes and edges in $G'$.
Then, a new representation is computed for each element $v \in V'$ by first computing relational representations. Given an element $v'$, each relation $(v', n') \in N(v')$ that $v'$ participates in, is converted to a \emph{relational representation}.

\begin{algorithm}
  \caption{The Relational Fusion Operator\label{alg:relationalFusion}}
  \begin{algorithmic}[1]\raggedright
    \Function{RelationalFusion$^k$}{$G'=(V', E')$, $\mathbf{H}^{(V', k-1)}$, $\mathbf{H}^{(E', k-1)}$}
      \State{\textbf{let} $\mathbf{H}^{(V', k)}$ be an arbitrary $|V'| \times d^{F^k}$ real feature matrix.}
      \ForAll{$v' \in V'$} 
          \State{$F_{v'} \gets \big\{$\par
          \hskip\algorithmicindent{} \hskip\algorithmicindent{} \hskip\algorithmicindent{}
          $\textsc{Fuse}^k(\mathbf{H}^{(V',k-1)}_{v'}, \mathbf{H}^{(E', k-1)}_{(v', n')}, \mathbf{H}^{(V', k-1)}_{n'}) 
                                     \mid n' \in N(v')\big\}$}
          \State{$\mathbf{H}^{(V', k)}_{v'} \gets \textsc{Aggregate}^k(F_{v'})$}
          \State{$\mathbf{H}^{(V', k)}_{v'} \gets \textsc{Normalize}^k(\mathbf{H}^{(V', k)}_{v'})$}
        \EndFor
        \State{\Return{$\mathbf{H}^{(V', k)}$}}
      \EndFunction
  \end{algorithmic}
\end{algorithm}

In \cref{alg:relationalFusion}, the relational representations at layer $k$ are computed by a fusion function $\textsc{Fuse}^k$. For each relation, $\textsc{Fuse}^k$ takes as input representations of the source $v'$ and target $n'$ of the relation, $\mathbf{H}^{(V', k-1)}_{v'}$ and $\mathbf{H}^{(V', k-1)}_{n'}$, respectively, along with a representation $\mathbf{H}^{(E', k-1)}_{(v', n')}$ describing their relation, and then it fuses them into a relational representation.
The relational representations are subsequently fed to a $\textsc{RelationalAggregate}^k$ function, that aggregates them into a single representation of $v'$.
Finally, the representation of $v'$ may optionally be normalized by invoking the $\textsc{Normalize}^k$ function, e.g., using $L_2$ normalization~\citep{graphsage}.
This latter step is particularly important if the relational aggregate has different scales across elements with different neighborhood sizes, e.g.\ if the aggregate is a sum.

Next, we discuss different choices for the fuse functions $\textsc{Fuse}^k$ and the relational aggregators $\textsc{Aggregate}^k$.

\subsection{Fusion Functions}
The fusion function is responsible for extracting the right information from each relation thereby allow an \gls{rfn} to create sharp boundaries at the edges of homophilic regions. The fusion function therefore plays an important role w.r.t.\ capturing volatile homophily.
We propose two fusion functions: $\textsc{AdditiveFuse}$ and $\textsc{InteractionalFuse}$.

$\textsc{AdditiveFuse}$ is among the simplest possible fusion functions.
Given input source, target, and relation representations $\mathbf{H}^{(V', k-1)}_{v'}$, $\mathbf{H}^{(V', k-1)}_{n'}$, and $\mathbf{H}^{(E', k-1)}_{(v', n')}$, \textsc{AdditiveFuse} transforms each of these sources individually using three trainable weight matrices:
\begin{multline}\label{eq:additiveFuse}
  \textsc{AdditiveFuse}^k(\mathbf{H}^{(V', k-1)}_{v'}, \mathbf{H}^{(V', k-1)}_{n'}, \mathbf{H}^{(E', k-1)}_{(v', n')}) = \\
  \sigma(
    \mathbf{H}^{(V', k-1)}_{v'}\mathbf{W}^s +
    \mathbf{H}^{(V', k-1)}_{n'}\mathbf{W}^t +
    \mathbf{H}^{(E', k-1)}_{(v', n')}\mathbf{W}^r +
    \mathbf{b})
\end{multline}
where $\sigma$ is an activation function, and where $\mathbf{W}^s \in \mathbb{R}^{d^s \times d^o}$, $\mathbf{W}^t \in \mathbb{R}^{d^t \times d^o}$, and $\mathbf{W}^r \in \mathbb{R}^{d^r \times d^o}$ are the source, target, and relation weights, respectively, and $\mathbf{b} \in \mathbb{R}^{1 \times d^o}$ is a bias term. 
Here, $d^o$ is the output dimensionality of the fusion function (and its parent relational fusion operator).

\cref{eq:additiveFuse} explicitly shows how the representations are transformed linearly and are summed.
It may be written more succinctly as 
\begin{multline*}
  \textsc{AdditiveFuse}^k(\mathbf{H}^{(V', k-1)}_{v'}, \mathbf{H}^{(V', k-1)}_{n'}, \mathbf{H}^{(E', k-1)}_{(v', n')}) = \\
  \sigma\big(\mathbf{H}^{(R, k-1)}_{(v', n')}\mathbf{W}^R + \mathbf{b}\big),
\end{multline*}
where $\mathbf{W}^R \in \mathbb{R}^{d^R \times d^o}$ is weight matrix with row dimensionality $d^R = d^s + d^t + d^r$, $\mathbf{H}^{(R, k-1)}_{(v', n')} = \mathbf{H}^{(V', k-1)}_{v'} \oplus \mathbf{H}^{(V', k-1)}_{n'} \oplus \mathbf{H}^{(E', k-1)}_{(v', n')}$ is a relational feature matrix, and $\oplus$ denotes vector concatenation.

$\textsc{AdditiveFuse}$ summarizes the relationship between $v'$ and $n'$, but does not explicitly model interactions between representations.
To that end, we propose an interactional fusion function
\begin{multline}
  \textsc{InteractionalFuse}^k(\mathbf{H}^{(V', k-1)}_{v'}, \mathbf{H}^{(V', k-1)}_{n'}, \mathbf{H}^{(E', k-1)}_{(v', n')}) = \\
  \sigma\big((\mathbf{H}^{(R, k-1)}\mathbf{W}^I \odot \mathbf{H}^{(R, k-1)})\mathbf{W}^R\big) + \mathbf{b},
\end{multline}
where $\mathbf{W}^I \in \mathbb{R}^{d^R \times d^R}$ is a trainable interaction weight matrix, $\odot$ denotes element-wise multiplication, and $\mathbf{b} \in \mathbb{R}^{1 \times d^o}$ is a bias term. 
When computing the term $\mathbf{H}^{(I, k)} = (\mathbf{H}^{(R, k-1)}\mathbf{W}^I) \odot \mathbf{H}^{(R, k-1)}$, the $i$th (for $1 \leq i \leq d^R$) value of vector $\mathbf{H}^{(I, k)}$ is
\begin{equation*}
  \mathbf{H}^{(I, k)}_i = \sum_{j=1}^{d^R} \mathbf{W}^I_{i, j}\mathbf{H}^{(R, k-1)}_i\mathbf{H}^{(R, k-1)}_j,
\end{equation*}
which enables $\textsc{InteractionalFuse}$ to capture and weigh interactions at a much finer granularity than $\textsc{AdditiveFuse}$.

InteractionalFuse offers improved modeling capacity over AdditiveFuse, enabling it to better address the challenge of volatile homophily, but at the cost of an increase in parameters that is quadratic in the number of input features to the fusion function.

\subsection{Relational Aggregators}
Many different \textsc{Aggregate} functions have been proposed in the literature, and many are directly compatible with the relational fusion layer.
For instance, \citet{graphsage} propose mean, LSTM, and max/mean pooling aggregators. 

Recently, aggregators based on attention mechanisms from the domain of natural language processing have appeared~\citep{gat}.
Such graph attention mechanisms allow a \gls{gcn} to filter out irrelevant or aberrant neighbors by weighing the contribution of each neighbor to the neighborhood aggregate separately.
This filtering property is highly desirable for road networks where even a single aberrant neighbor can contribute significant noise to an aggregate due to the low density of road networks.
In addition, it may help the network distinguish adjacent regions from each other and thereby improve predictive performance in the face of volatile homophily.

Current graph attention mechanisms rely on a common transformation of each neighbor thus rendering them incompatible with \glspl{rfn} that rely on the context-dependent neighbor transformations performed by the \textsc{Fuse} function at each relational fusion layer.
We therefore propose an attentional aggregator that is compatible with our proposed \glspl{rfn}.


\paragraph{Attentional Aggregator}\label{sec:attentionalAggregator}
The attentional aggregator we propose computes a weighted mean over the relational representations.
Formally, the attentional aggregator computes the $\textsc{Aggregate}$ in \cref{alg:relationalFusion} as
\begin{equation}\label{eq:attentionalAggregator}
  \textsc{Aggregate}(F_v) = \sum_{\mathbf{f}_n \in F_v} A(v,n)\mathbf{f}_n,
\end{equation}
where $F_v = \{\mathbf{f}_n \mid n \in N(v)\}$ is the set of fused relational representations of node $v$'s relations to each of its neighbors $n \in N(v)$ computed at line $4$ in \cref{alg:relationalFusion}.
Furthermore, $A$ is an attention function, and $A(v, n)$ is the \emph{attention weight} that determines the contribution of each neighbor $n$ to the neighborhood aggregate of node $v$.

An attention weight $A(v', n')$ in \cref{eq:attentionalAggregator} depends on the relationship between a node $v'$ and its neighbor $n'$ in the input graph $G'$ to \cref{alg:relationalFusion}.
This relationship is described by the relational feature matrix $\mathbf{H}^{(R, k-1)}_{(v', n')} \in \mathbb{R}^{d^R}$ and the attention weight is computed as
\begin{equation}\label{eq:attentionWeight}
  A(v', n') = \frac{
    \exp\big(C(\mathbf{H}^{(R, k-1)}_{(v', n')}) \big)}
  {\sum_{n'' \in N(v')} \exp\big( C(\mathbf{H}^{(R, k-1)}_{(v', n'')}) \big)},
\end{equation}
where $C \colon \mathbb{R}^{d^R} \xrightarrow{} \mathbb{R}$ is an \emph{attention coefficient} function
\begin{equation}
  C(\mathbf{H}^{(R, k-1)}_{(v', n')}) =
      \sigma(\mathbf{H}^{(R, k-1)}_{(v', n')} \mathbf{W}^C),
\end{equation}
where $\mathbf{W^C} \in \mathbb{R}^{d^R}$ is a weight matrix and $\sigma$ is an activation function.
This corresponds to computing the softmax over the attention coefficients of all neighbors $n$ of $v$.
All attention weights sum to one, i.e., $\sum_{n \in N(v)} A(v, n) = 1$, thus making \cref{eq:attentionalAggregator} a weighted mean.
In other words, each neighbor's contribution to the mean is regulated by an attention weight, thus allowing an RFN to reduce the influence of (or completely ignore) aberrant neighbors that would otherwise contribute significant noise to the neighborhood aggregate.

\subsection{Forward Propagation}
\begin{algorithm}
  \caption{Forward Propagation Algorithm\label{alg:forward-propagation}}
  \begin{algorithmic}[1]\raggedright
    \Function{ForwardPropagation}{$\mathbf{X}^V$, $\mathbf{X}^E$, $\mathbf{X}^B$}
      \State{\textbf{let} $\mathbf{H}^{(V, 0)} = \mathbf{X}^V$, $\mathbf{H}^{(E, 0)} = \mathbf{X}^{E}$, and $\mathbf{H}^{(B, 0)} = \mathbf{X}^{B}$}
      \For{$k=1$ to $K$}
        \State{$\mathbf{H}^{(V, k)} \gets \textsc{RelationalFusion}^k(
          G^P, \mathbf{H}^{(V, k-1)}, \mathbf{H}^{(E, k-1)})$}
        \State{$\mathbf{H}^{(B', k-1)} \gets \textsc{Join}(\mathbf{H}^{(V, k-1)}, \mathbf{H}^{(B, k-1)})$}
        \State{$\mathbf{H}^{(E, k)} \gets \textsc{RelationalFusion}^k(
          G^D, \mathbf{H}^{(E, k-1)}, \mathbf{H}^{(B', k-1)})$}
        \State{$\mathbf{H}^{(B, k)} \gets \textsc{FeedForward}^k(\mathbf{H}^{(B, k-1)})$}
        \EndFor
        \State{\Return{$\mathbf{H}^{(V, K)}$, $\mathbf{H}^{(E, K)}$, $\mathbf{H}^{(B, K)}$}}
      \EndFunction
    \Function{Join}{$\mathbf{H}^{V}$, $\mathbf{H}^{E}$}
    \ForAll{$\big((u, v), (v, w)\big) \in B$}
      \State{$\mathbf{H}^{B'}_{\big( (u, v), (v, w) \big)} \gets \mathbf{H}^B_{\big( (u, v), (v, w) \big)} \oplus \mathbf{H}^V_v$}
    \EndFor
    \State{\Return{$\mathbf{H}^{B'}$}}
    \EndFunction
  \end{algorithmic}
\end{algorithm}

With the relational fusion operator and its components defined in Sections 4.3 to 4.5, we proceed to explain how forward propagation is performed through the constituent relational fusion layers (introduced in Section 4.2) of an RFN. The forward propagation algorithm is shown in Algorithm 2.
Starting from the input encoding of node, edge, and between-edge attributes, each layer $k$ transforms the node, edge, and between-edge representations emitted from the previous layer.
The node representations are transformed using node-relational fusion.
This is done by invoking the $\textsc{RelationalFusion}$ function in line~$4$ with the primal graph, and the node and edge representations from the previous layer as input.

Edge-relational fusion is performed in lines~$5\text{-}6$ to transform the edge representations from the previous layer.
In line~$5$, the node and between-edge representations from the previous layer are joined using the \textsc{Join} function (defined on lines~$9\text{-}12$) to capture the information from both sources that describe the relationships between two edges.
On line~$6$, $\textsc{RelationalFusion}$ is invoked again, now with the dual graph (indicating relationships between edges), and the edge representations from the previous layer.
The last input is the joined node and between-edge representations from the previous layer.
Finally, the between-edge representations are transformed using a single feed-forward operator in line~$7$.

Notice that the number of layers in an \gls{rfn} determines which nodes, edges, and between-edges influence the output representations.
With $K$ layers, the relational fusion operators in the node- and edge-relational views aggregate information up to a distance $K$ from nodes in the primal graph and edges in the dual graph, respectively.
Notice also that the forward propagation algorithm in \cref{alg:forward-propagation} produces three outputs: node, edge, and between-edge predictions.
This enables the relational fusion network to be jointly optimized for node, edge, and between-edge predictions, e.g., when the network is operating in a multi-task learning setting.
If only one or two outputs are desired, the superfluous operations in the last layer can be ''switched off'' to save computational resources.
For instance, propagation of node and between-edges can be ''switched off'' by replacing line~$4$ and line~$7$ with $\mathbf{H}^{(V, k)} = \mathbf{H}^{(V, k-1)}$ and $\mathbf{H}^{(B, k)} = \mathbf{H}^{(B, k-1)}$, respectively, when $k = K$.

\section{Experimental Evaluation}\label{sec:experiments}
To investigate the generality of our method, we evaluate it on two tasks using the road network of the Danish municipality of Aalborg: driving speed estimation and speed limit classification. These tasks represent a regression task and a classification task, respectively.

\subsection{Data Set}
We extract the spatial representation of the Danish municipality of Aalborg from \gls{osm}~\citep{osm}, and convert it to primal and dual graph representations as described in \cref{sec:road-network-modelling}.
We derive node attributes using a zone map from the Danish Business Authority\footnote{https://danishbusinessauthority.dk/plansystemdk}.
The map categorizes regions in Denmark as city zones, rural zones, and summer cottage zones. We use a node attribute for each zone category to indicate the category of an intersection.
In the \gls{osm} data, each road segment is categorized as one of nine road categories indicating their importance.
We use the road segment category and length as edge attributes.
As between-edge attributes, we use turn angles (between $0$ and $180$ degrees) relative to the driving direction between adjacent road segments as well as the turn directions, i.e., right-turn, left-turn, U-turn, or straight-ahead.
This separation of turn angle and turn direction resulted in superior results during early prototyping of our method.

\subsubsection{Attribute Encoding}
We encode each of the node attributes using a binary value, $0$ or $1$, thus obtaining node encodings with $3$ features per node.
Edge attributes are encoded as follows. Road categories are one-hot encoded, and road segment lengths are encoded as continuous values, yielding a total of $10$ features per edge.
We encode between-edge attributes by one-hot encoding the turn direction and representing the turn angle (between $0$ and $180$ degrees) using a continuous value, yielding an encoding with $5$ features per between-edge.

Conventional \glspl{gcn} can only use a single source of attributes.
The \glspl{gcn} we use in our experiments are therefore run on the dual graph and leverages the $10$ edge features.
However, to achieve more fair comparisons, we concatenate the source and target node features to the encodings of the edge attributes. This yields a final edge encoding with $16$ features per edge that we use throughout our experiments.
This allows the conventional \glspl{gcn} to also leverage node attributes.

Finally, we use min-max normalization s.t.\ all feature values are in the $[0; 1]$ range.
This normalization ensures that all features are on the same scale, thus making training with the gradient-based optimization algorithm we use in our experiments more stable.

\subsubsection{Driving Speeds}
For the driving speed estimation tasks, we extract driving speeds from a set of $336\,253$ vehicle \gls{gps} trajectories with a \gls{gps} sampling frequency of 1Hz. The trajectories have been collected by a Danish insurance company between 2012--01--01 and 2014--12--31 and processed by \citet{driving_speeds}.
Each vehicle trajectory consists of a number of records that record time and position. Each record is converted to a driving speed and mapmatched to a road segment~\citep{driving_speeds}, i.e., an edge in the primal graph.
We count each such record as an observation in our data set.

We split the data set into training, validation, and test sets. The training, validation, and test set contains observations derived from trajectories that started between 2012--01--01 and 2013--06--31, 2013--07--01 and 2013--12--31, and 2014--01--01 and 2014--12--31, respectively. 
The driving speeds are highly time-dependent.
For the purposes of these experiments, we therefore exclude data in typical Danish peak hours between 7~a.m.\ and 9~a.m.\ in the morning and between 3~p.m.\ and 5~p.m.\ in the afternoon to reduce the variance caused by time-varying traffic conditions.

\subsubsection{Speed Limits}
For the speed limit classification task, we use speed limits collected from the \gls{osm} data and additional speed limits collected from the Danish municipality of Aalborg~\citep{workshop}.
This dataset is quite imbalanced, e.g., there are $15\,582$ road segments that have a speed limit of $50$~kmh and just $36$ road segments that have a speed limit of $130$~kmh.
In addition, there is a substantial geographical skew in the date since much of the data is crowd-sourced: most speed limits in the dataset are within highly populous regions such as city centers.
We split the speed limits such that $50\%$, $25\%$, and $25\%$ are used for training validation, and testing, respectively. 

See \cref{tab:dataset-summary} for detailed statistics on these datasets.

\begin{table}[h]
  \centering
  \caption{Statistics of the datasets used in our experiments.\label{tab:dataset-summary}}
  \footnotesize
  \begin{tabular}{llr}
    \toprule
    \multicolumn{3}{l}{\emph{Road Network Characteristics}}\\
    & No.\ of Nodes & $16\,294$ \\
    & No.\ of Edges & $35\,947$ \\
    & No.\ of Between-Edges & $94\,718$ \\
    & No.\ of Node Features & $3$ \\
    & No.\ of Edge Features & $16$ \\
    & No.\ of Between-Edge Features & $5$ \\
    \bottomrule
  \end{tabular}
  
  \vspace*{0.25cm}
  \begin{tabular}{lrr}
    \toprule
    & \multicolumn{1}{r}{\emph{Driving Speeds}} & \multicolumn{1}{r}{\emph{Speed Limits}} \\
    \cmidrule(lr){2-2} \cmidrule(lr){3-3}
     Training Set Size & $4\,031\,964$ & $9\,710$ \\
     Validation Set Size & $1\,831\,775$ & $4\,534$ \\
     Test Set Size & $2\,811\,860$ & $5\,266$ \\
     Total Data Set Size & $8\,675\,599$ & $19\,510$ \\
    \bottomrule
  \end{tabular}
\end{table}

\subsection{Algorithms}
We use the following baselines for comparison in the experiments:
\begin{itemize}
  \item \emph{Grouping Estimator}: A simple baseline that groups all road segments depending on their road category and on whether they are within a city zone or not. At prediction time, the algorithm outputs the mean (for regression) or mode (for classification) label of the group a particular road segment belongs to.
  \item \emph{MLP}: A regular multi-layer perceptron that performs predictions independent of adjacent road segments by using only the edge encodings as input.
  \item \emph{GraphSAGE}: The Max-Pooling variant of GraphSAGE, which achieved the best results in the authors' experiments~\citep{graphsage}.
  \item \emph{GAT}: The graph attention network by \citet{gat}.
\end{itemize}

To analyze the importance and impact of the different combinations of relational aggregators and fusion functions, we use four variants of the relational fusion network against these baselines:
\begin{itemize}
  \item \emph{RFN-Attentional+Interactional}: A relational fusion network using the \emph{attentional} aggregator with \emph{interactional} fusion.
  \item \emph{RFN-Attentional+Additive}: A relational fusion network using the \emph{attentional} aggregator with \emph{additive} fusion.
  \item \emph{RFN-NonAttentional+Interactional}: A relational fusion network using a \emph{non-attentional} aggregator with \emph{interactional} fusion. The non-attentional aggregator uses a constant attention coefficient across all neighbors when computing the relational aggregate corresponding to an arithmetic mean.
  \item \emph{RFN-MI}: A relational fusion network using a \emph{non-attentional} aggregator with \emph{additive} fusion.
\end{itemize}

\subsection{Experimental Setup}
We implement all algorithms based on neural networks using the MXNet\footnote{https://mxnet.incubator.apache.org/} deep learning library.
We make our implementation of the relational fusion networks publicly available online~\footnote{\url{https://github.com/TobiasSkovgaardJepsen/relational-fusion-networks}}.

\subsubsection{Hyperparameter Selection}
We use an architecture with two layers for all neural network algorithms.
For the \glspl{rfn}, we use $L_2$ normalization~\citep{graphsage} on some layers, i.e., $\textsc{Normalize}(\mathbf{h}) = \frac{\mathbf{h}}{|\mathbf{h}|_2}$ for a feature vector $h \in \mathbb{R}^d$ where $|\cdot|_2$ is the $L_2$ norm, to ensure that the output of each layer is between $-1$ and $1$.
For driving speed estimation, we use $L_2$ normalization on the first layer. For speed limit classification, we found that training became more stable when using $L_2$ normalization on both layers.
In addition, we omit the $L_2$ normalization on the last layer of GraphSAGE for driving speed estimation; otherwise, the output cannot exceed a value of one.

We use the ReLU~\citep{relu} and softmax activation functions on the last layer for driving speed estimation and speed limit classification, respectively.
The ReLU function ensures outputs of the model are non-negative s.t.\ no model can estimate negative speeds.
Following the experimental setup of the authors of GraphSAGE~\citep{graphsage}, we use the ReLU activation function for the GraphSAGE pooling network.
For all attention coefficient networks in the GAT algorithm and relational fusion network, we use the LeakyReLU activation function with a negative input slope of $\alpha=0.2$, like the GAT authors~\citep{gat}.
For all other activation functions, we use the \gls{elu} activation function.
We select the remaining hyperparameters using a grid search and select the configuration with the best predictive performance on the validation set.

Based on preliminary experiments, we explore different learning rates $\lambda \in \{0.1, 0.01, 0.001\}$ in the grid search.
For GraphSAGE, GAT, and all RFN we explored $d \in \{32, 64, 128\}$ output dimensionalities of the first layer .
The MLP uses considerably fewer parameters than the other algorithms, and we therefore explore larger hidden layer sizes $d \in \{128, 256, 512\}$ for a fair comparison.

The GraphSAGE and GAT algorithms have additional hyperparameters.
GraphSAGE uses a pooling network to perform neighborhood aggregation. For each layer in GraphSAGE, we set the output dimension of the pooling network to be double the output dimension of the layer in accordance with the authors' experiments~\citep{graphsage}.

The GAT algorithm uses multiple attention heads that each output $d$ features at the first hidden layer.
These features are subsequently concatenated yielding an output dimensionality $h \cdot d$ where $h$ is the number of attention heads.
Based on the work of the authors~\citep{gat}, we explore different values of $h \in \{1, 2, 4, 8\}$ during the grid search.
However, due to the concatenation, large values of $h$ combined with large values of $d$ make the GAT network very time-consuming to train.
We therefore budget these parameters s.t.\ $h \cdot d \leq 256$ corresponding to, e.g., a GAT network with $64$ output units from $4$ attention heads.

\subsubsection{Model Training and Evaluation}
We initialize the neural network weights using Xavier initialization~\citep{xavier} and train the models using the ADAM optimizer~\citep{adam} in batches of $256$ segments.
In preliminary experiments, we observed that all models converged within $20$ and $30$ epochs for driving speed estimation and speed limit classification, respectively.
We therefore use these values for training.

In order to speed up training, we train on a sub-network induced by the each batch (similar to the mini-batch training approach of \citet{graphsage}) s.t.\ the graph convolutional models only perform the computationally expensive graph convolutions on the relevant parts of the network.
However, to avoid the computational overhead of generating the sub-network for each batch in each epoch, we pre-compute the batches and shuffle the batches during each epoch.
In order to ensure that the mini-batches provide good approximations of the global gradient, we compute the batches in a stratified manner.
In the case of the driving speed estimation task, we select road segments for each batch s.t.\ the distribution of road categories in the batch is similar to the distribution of the entire training set.

\paragraph{Driving Speed Estimation}
In this experiment, we aim to estimate the mean driving speed on each road segment in a road network.
For the driving speed estimation task we evaluate each model by measuring the \gls{mae} between the mean recorded speed of a segment and the model output.
Let $D=\bigcup_{i=1}^{N}\{(s_i, Y_i)\}$ be a dataset, e.g., the training set, where $(s_i, Y_i)$ is a unique entry for road segment $s_i \in E$ and $Y_i = {y_{i, 1}, \dots, y_{i, j} }$ is a set of recorded speeds on road segment $s_i$.
Formally, we measure the error of each model as $\text{MAE} = \frac{1}{N}\sum_{(s, Y) \in D} |\hat{y} - \bar{Y}|$ where $\hat{y}$ is the estimated driving speed of segment $s$ of the model and $\bar{Y}=\sum_{y \in Y} \frac{y}{|Y|}$ is the mean recorded speed for a segment $s$.
$\bar{Y_i}$ is not representative of the population mean of a road segment $s_i$ if it contains very few speed records.
We therefore remove entries $(s_i, Y_i)$ from $D$ if there are fewer than ten recorded speeds in $Y_i$ when measuring \gls{mae}.

The recorded driving speeds are heavily concentrated on a few popular road segments.
We therefore weigh the contribution of each recorded speed to the loss s.t.\ each road segment contribute evenly to the loss independent of their frequency in the dataset.
Formally, we minimize the average (over segments) \gls{mse} loss of a model: $\textsc{AMSE-Loss} = \frac{1}{N}\sum_{(s, Y) \in D} \sum_{y \in Y} \frac{(\hat{y} - y)^2}{|Y|}$ where $\hat{y}$ is the estimated driving speed of segment $s$ of the model.

\paragraph{Speed Limit Classification}
We follow the methodology of \citet{workshop} and use random over-sampling with replacement when training the model to address the large imbalance in speed limit frequencies.
This ensures that all speed limits occur with equal frequency during training.
We train the model to minimize the categorical cross entropy on the over-sampled training set and measure model performance using the macro $F_1$ score which punishes poor performance equally on all speed limits, irrespective of their frequency.

During training, we found that all algorithms were prone to overfitting on the training set.
In addition, the model decision boundaries between speed limits are very sensitive during training causing the macro $F_1$ score on the validation set to be highly unstable.
We therefore use a variant of early stopping to regularize the model: we store the model after each training epoch and select the version of the model with the highest macro $F_1$ on the validation set.
Finally, we restore this model version for final evaluation on the test set.

\subsection{Results}
We run ten experiments for each model and report the mean performance with standard deviations in \cref{tab:results}.
The GAT algorithm can become unstable during training~\citep{gat} and we observed this phenomenon on one out of the ten runs on the driving speed estimation task.
The algorithm did not converge on this run and is therefore excluded from the results shown in \cref{tab:results}.
Note that when reading \cref{tab:results}, low values and high values are desirable for driving speed estimation and speed limit classification, respectively.

\begin{table}[]
  \centering
  \caption{Algorithm performance on Driving Speed Estimation (DSE) and Speed Limit Classification (SLC).\label{tab:results}}
  \small
  \begin{tabular}{lll}
    \toprule
    \textit{Algorithm}             & \textit{DSE}   & \textit{SLC}  \\
    \midrule
    Grouping Predictor             & $11.026$         & $0.356$         \\
    MLP                            & $10.160 \pm 0.119$ & $0.443 \pm 0.027$ \\
    GraphSAGE                      & $8.960 \pm 0.115$  & $0.432 \pm 0.014$ \\
    GAT                            & $9.548 \pm 0.151$ & $0.442 \pm 0.018$ \\
    \midrule
    RFN-Attentional+Interactional  & $\mathbf{6.797 \pm 0.124}$ & $\mathbf{0.535 \pm 0.014}$   \\
    RFN-NonAttentional+Interactional & $6.911 \pm 0.080$  & $0.507 \pm 0.012$ \\
    RFN-Attentional+Additive     & $7.440 \pm 0.133$  &     $0.518 \pm 0.022$         \\
    RFN-NonAttentional+Additive    & $7.685 \pm 0.189$  &     $0.500 \pm 0.011$          \\
    \bottomrule
  \end{tabular}
\end{table}

As shown in \cref{tab:results}, our relational fusion network variants outperform all baselines on both driving speed estimation and speed limit classification.
The best \gls{rfn} variant outperforms the state-of-the-art graph convolutional approaches, i.e., GraphSAGE and GAT, by $32\%$ and $40\%$, respectively, on the driving speed estimation task.
On the speed limit classification task, the best \gls{rfn} variant outperforms GraphSAGE and GAT by $24\%$ and $21\%$, respectively.

\cref{tab:results} shows that the different proposed variants are quite close in performance, but the attentional variants are superior to their non-attentional counterparts that use the same fusion function. In addition, the interactional fusion function appears to be strictly better than the additive fusion function.
Interestingly, GraphSAGE and GAT fail to outperform MLP on the speed limit classification task.
The MLP classifies road segments independent of any adjacent road segments.
Unlike the \gls{rfn} variants, it appears that GraphSAGE and GAT are unable to effectively leverage the information from adjacent road segments.
This supports our discussion in \cref{sec:introduction} of the problems with direct inheritance during neighborhood aggregation in the context of road networks.

\subsection{Case Study: Capturing Volatile Homophily}
As discussed in \cref{sec:introduction}, road networks exhibit volatile homophily characterized by abrupt changes in network characteristics over short network distances.
We therefore perform a case study to assess \glspl{rfn}' capacity for handling volatile homophily compared to the two \gls{gcn} baselines, GAT and GraphSAGE.

\begin{figure}
  \centering
  \begin{subfigure}{0.37875\columnwidth}
    \centering
    \includegraphics[width=\textwidth, trim={23cm, 0cm, 25cm, 0cm}, clip]{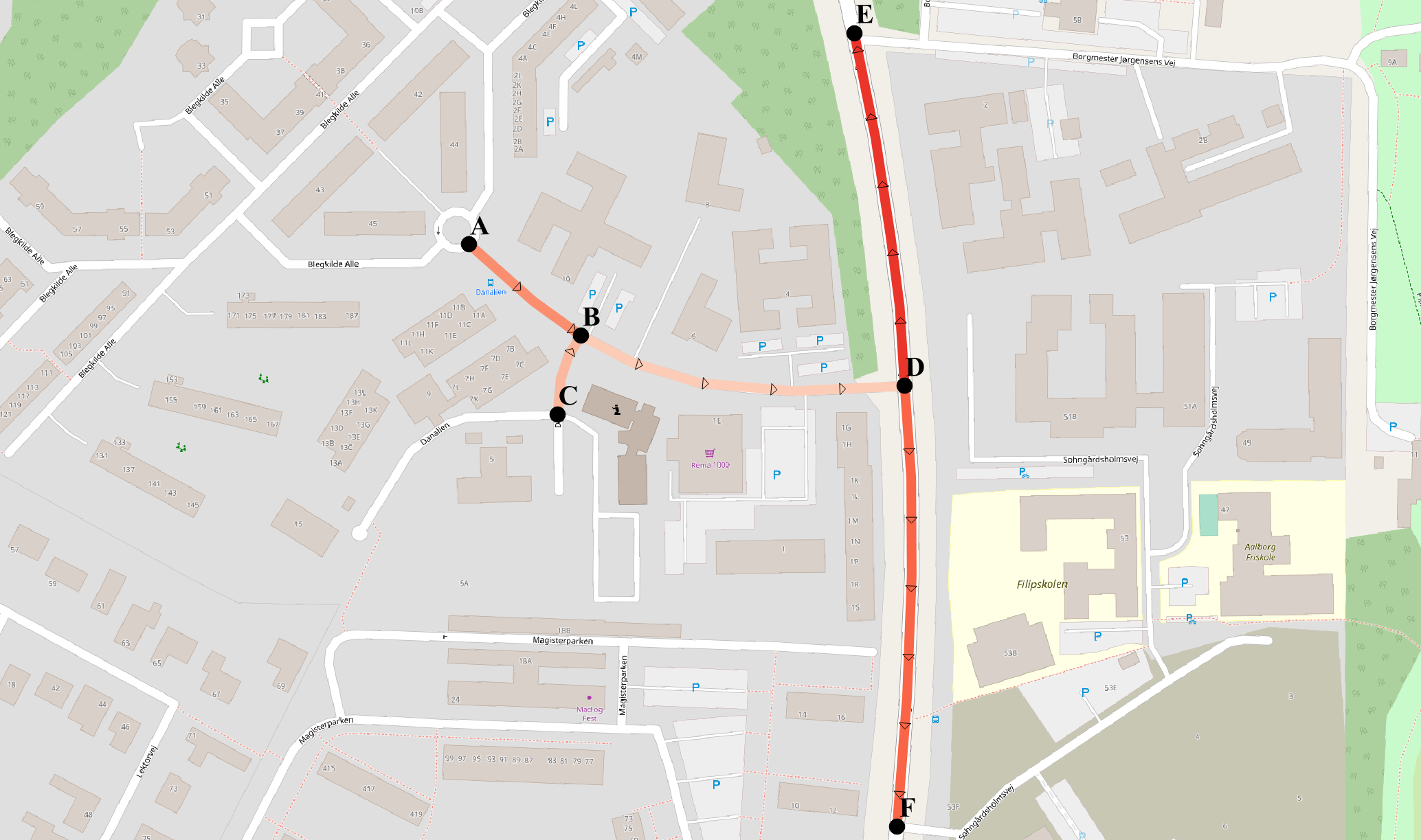}
    \caption{Danalien.\label{fig:volatile-homophily-a}}
  \end{subfigure}
  \begin{subfigure}{0.61125\columnwidth}
    \centering
    \includegraphics[width=\textwidth, trim={1cm, 0cm, 1cm, 0cm}, clip]{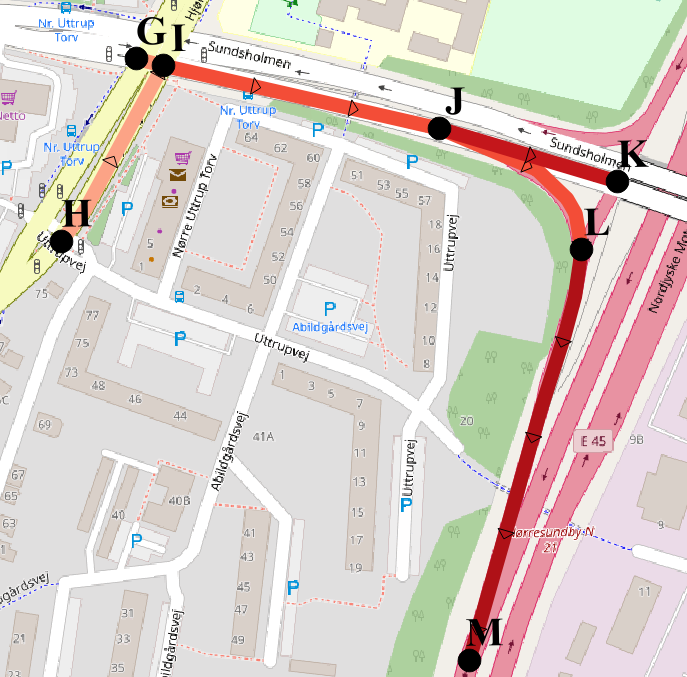}
    \caption{Nordjydske Motorvej.\label{fig:volatile-homophily-b}}
  \end{subfigure}
  \caption{Ground truth driving speeds of two regions in Aalborg, Denmark. Triangles indicate direction, black dots mark nodes, and color indicates speed: the darker, the faster.\label{fig:volatile-homophily-examples}}
\end{figure}

\subsubsection{Cases}
We select the two regions shown in \cref{fig:volatile-homophily-examples}.
First, we examine the case in \cref{fig:volatile-homophily-a} which shows an expanded view of the three-way intersection from \cref{fig:compelling-example}.
The segments DE and DF comprise a busy transportation region that connects different residential regions.
The segments AB, CB, and BD are exits from one such residential region.
This case is characterized by the sharp boundary between these two regions, as indicated by the colors on the figure.
The AB and BD segments form part of the street Danalien.
We therefore refer to this case as Danalien.

Next, we examine the boundary region between the ordinary city segments and the motorway segments shown in \cref{fig:volatile-homophily-b}.
Here, segments GI, HI, IJ, and JK are city roads, and JL and LM are motorway segments.
This case is characterized by the steep increase in driving speed when moving from node G or H to node K or M.
The segments JL and LM form part of the motorway Nordjydske Motorvej, and we therefore refer to this case as Nordjydske Motorvej.

For both \cref{fig:volatile-homophily-a} and \cref{fig:volatile-homophily-b}, we report the ground truth driving speed and predicted driving speeds of the best performing RFN-AA-I, GAT, and GraphSAGE models on the validation set in \cref{tab:volatile-homophily-speeds}.
We also report the \gls{mae} score for the subnetwork in each case.

\COMMENT{
\begin{table}
  \caption{Predicted and ground truth driving speeds from the subnetworks shown in \cref{fig:volatile-homophily-a} and \cref{fig:volatile-homophily-b}.
           Scores are calculated for each subnetwork.\label{tab:danalien-speeds}}
  \begin{tabular}{lrrrr}
    \toprule
    \emph{Segment} & \emph{Ground Truth} & \emph{RFN-AA-I} & \emph{GAT} & \emph{GraphSAGE} \\ 
    \midrule
    AB & $31.9174099257860$ & $22.3250$ & $26.8734$ & $27.1498$ \\
    BD & $17.7430390133737$ & $20.9658$ & $20.8766$ & $25.4401$ \\
    CB & $24.1101507406968$ & $17.3884$ & $27.2529$ & $27.0424$ \\
    DE & $51.4384779863299$ & $56.8684$ & $47.4659$ & $45.4440$ \\
    DF & $41.9618431580746$ & $56.0193$ & $43.9899$ & $45.1932$ \\
    \emph{Score} & $0$	& $7.804860102$	& $\mathbf{3.464191}$ & $4.924551$ \\
    \midrule
    GI & $32.9527931598363$ & $27.6162$ & $36.7828$ & $21.1801$ \\
    HI & $25.3582789022965$ & $36.5841$ & $38.9427$ & $39.8348$ \\
    IJ & $47.2636261741656$ & $40.2105$ & $49.5860$ & $34.8590$ \\
    JK & $62.5301182895355$ & $48.0588$ & $48.1628$ & $44.0977$ \\
    JL & $45.7028056657979$ & $42.0992$ & $67.5647$ & $53.8469$ \\
    LM & $66.9090825927627$ & $59.6024$ & $56.5644$ & $63.7620$ \\
    \emph{Score} & $0$ & $\mathbf{8.338092877}$ & $11.19320288$ & $13.04607061$ \\
    \bottomrule
  \end{tabular}
\end{table}
}
\begin{table}
  \caption{Predicted and ground truth driving speeds from the subnetworks shown in \cref{fig:volatile-homophily-a} and \cref{fig:volatile-homophily-b}.
           Scores are calculated for each subnetwork.\label{tab:volatile-homophily-speeds}}
   \footnotesize
  \begin{tabular}{llrrrr}
    \toprule
    \emph{Segment} & \emph{Ground Truth} & \emph{RFN-AA-I} & \emph{GAT} & \emph{GraphSAGE} \\ 
    \midrule
    AB & $31.92$ & $22.33$ & $26.87$ & $27.15$ \\
    BD & $17.74$ & $20.97$ & $20.88$ & $25.44$ \\
    CB & $24.11$ & $17.39$ & $27.25$ & $27.04$ \\
    DE & $51.44$ & $56.87$ & $47.47$ & $45.44$ \\
    DF & $41.96$ & $56.02$ & $43.99$ & $45.19$ \\
    \midrule
    \emph{Score} & $0$	& $7.80$	& $\mathbf{3.46}$ & $4.92$ \\
    \toprule
    GI & $32.95$ & $27.62$ & $36.78$ & $21.18$ \\
    HI & $25.36$ & $36.58$ & $38.94$ & $39.83$ \\
    IJ & $47.26$ & $40.21$ & $49.59$ & $34.86$ \\
    JK & $62.53$ & $48.06$ & $48.16$ & $44.10$ \\
    JL & $45.70$ & $42.10$ & $67.56$ & $53.84$ \\
    LM & $66.91$ & $59.60$ & $56.56$ & $63.76$ \\
    \midrule
    \emph{Score} & $0$ & $\mathbf{4.00}$ & $7.25$ & $5.27$ \\
    \bottomrule
  \end{tabular}
\end{table}

\subsubsection{Danalien}
\begin{table}
  \centering
  \footnotesize
  \caption{Mean and standard deviations of driving speeds within each region of the Danalien case.\label{tab:danalien-summary}}
  \begin{tabular}{lrrrr}
    \toprule
    \emph{Region} & \emph{Ground Truth} & \emph{RFN-AA-I} & \emph{GAT} & \emph{GraphSAGE} \\
    \midrule
    Residential & $24.59 \pm 7.10$ & $20.23 \pm 2.55$ & $25.00 \pm 3.58$ & $26.54 \pm 0.96$ \\
    Transportation & $46.70 \pm 6.70$ & $56.44 \pm 0.60$ & $45.73 \pm 2.46$ & $45.32 \pm 0.18$ \\
    \bottomrule
  \end{tabular}
\end{table}

We report the mean and standard deviations predicted driving speeds within the residential region and transportation region of the Danalien case \cref{tab:danalien-summary}.
The statistics for the ground truth are included in the table for comparison.
As shown in the table, all model assign quite different different driving speeds to within each region.
Thus, it would appear that all models find a boundary between the residential and the transportation region.

We examine the sharpness of this boundary by comparing the mean predictions within the residential region and the transportation region.
We find that the RFN-AA-I model is capable of making a much sharper difference with an absolute difference between the region means of $36.21$.
In comparison, the GAT and GraphSAGE models has absolute differences of $20.73$ and $18.78$, respectively, which is consistent with the underlying "smooth" homophily assumption of these two models.

In the Danalien case, the RFN-AA-I model performs worse than the GAT and GraphSAGE models, as indicated by the scores in \cref{tab:volatile-homophily-speeds}.
However, as documented in \cref{tab:results}, the sharp boundaries of the RFN-AA-I algorithm are generally advantageous.

\begin{table*}[]
  \centering
  \caption{Attention weights for when RFN-AA-I model predictions on segments (a) BD, (b) IJ, and (c) JL.}
  \begin{subtable}{0.4\textwidth}
    \centering
  \caption{Segment BD.\label{tab:attention-weights-cd}}
  \footnotesize
  \begin{tabular}{crrrrrrr}
    \toprule
    & \multicolumn{3}{c}{\emph{In-Neighbors}} & \multicolumn{3}{c}{\emph{Out-Neighbors}} \\
    \cmidrule(lr){2-4} \cmidrule(lr){5-7}
    \emph{Layer} & \multicolumn{1}{c}{AB} & \multicolumn{1}{c}{CB}& \multicolumn{1}{c}{DB}
                                     & \multicolumn{1}{c}{BB} & \multicolumn{1}{c}{DE} & \multicolumn{1}{c}{DF} \\
    \midrule
    $1$ & $17.9\%$ & $17.1\%$ & $13.5\%$ & $13.5\%$ & $17.3\%$ & $20.8\%$ \\
    $2$ & $14.7\%$ & $9.0\%$ & $6.4\%$ & $6.4\%$ & $32.6\%$  & $30.9\%$ \\
    \bottomrule
\end{tabular}
  \end{subtable}
  \begin{subtable}{0.3\textwidth}
    \centering
  \caption{Segment IJ.\label{tab:attention-weights-ij}}
  \footnotesize
  \begin{tabular}{crrrrr}
    \toprule
    & \multicolumn{2}{c}{\emph{In-Neighbors}} & \multicolumn{2}{c}{\emph{Out-Neighbors}} \\
    \cmidrule(lr){2-3} \cmidrule(lr){4-5}
    \emph{Layer} & \multicolumn{1}{c}{GI} & \multicolumn{1}{c}{HI}
                 & \multicolumn{1}{c}{JK} & \multicolumn{1}{c}{JL} \\
    \midrule
    $1$ & $24.9\%$ & $22.2\%$ & $30.5\%$ & $22.4\%$ \\
    $2$ & $19.3\%$ & $23.2\%$ & $31.3\%$ & $26.1\%$ \\
    \bottomrule
\end{tabular}
  \end{subtable}
  \begin{subtable}{0.26\textwidth}
    \centering
  \caption{Segment JL.\label{tab:attention-weights-jl}}
  \footnotesize
  \begin{tabular}{ccc}
    \toprule
    & \multicolumn{1}{c}{\emph{In-Neighbors}} & \multicolumn{1}{c}{\emph{Out-Neighbors}} \\
    \cmidrule(lr){2-2} \cmidrule(lr){3-3}
    \emph{Layer} & \multicolumn{1}{c}{IJ} 
                                     & \multicolumn{1}{c}{LM} \\
    \midrule
    $1$ & $73.3\%$ & $26.7\%$ \\
    $2$ & $73.5\%$ & $26.5\%$ \\
    \bottomrule
\end{tabular}
  \end{subtable}
\end{table*}

\subsubsection{Nordjydske Motorvej}
\begin{figure}
  \vspace*{-0.5cm}
  \centering{
  \scalebox{0.55}{\input{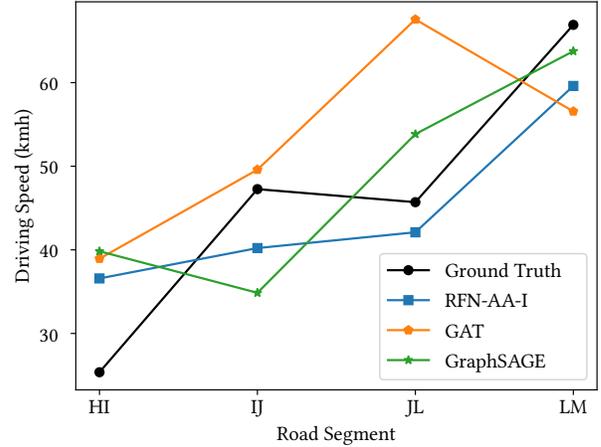}}
}
  \vspace*{-0.75cm}
    \caption{Driving speeds of different road segments along the path H-I-J-L-M.\label{fig:path-plot}}
  \vspace*{-0.75cm}
\end{figure}

Unlike the Danalien case, the Nordjydske Motorvej case shown in \cref{fig:volatile-homophily-b} does not have two clearly defined areas.
Instead, there is a continuous, but steep change in driving speeds when traveling from nodes G or H to nodes K or M.
We therefore analyze model behavior from the perspective of the path H-I-J-L-M.
We plot the ground truth and predicted driving speeds at each road segment along the path in \cref{fig:path-plot} using the values shown in \cref{tab:volatile-homophily-speeds}.

The H-I-J-L-M path has a complex curve as shown in \cref{fig:path-plot}.
As illustrated by the figure, there is a plateau in driving speeds at segment IJ and JL in the path.
All models substantially overestimate the driving speed on segment HI, but from thereon the RFN-AA-I more closely matches the ground truth curve.
Interestingly, both the GraphSAGE model and GAT model fail to capture the plateau, whereas the RFN-AA-I model has only a small positive slope between segments IJ and JL.\ In addition, the GAT model again fails to predict the increase on the last segment in the path.

We suspect that the GraphSAGE and GAT model fail to capture the plateau in \cref{fig:path-plot} because they directly inherit the representation of the motorway segment LM which has a substantially higher driving speed as shown in \cref{tab:volatile-homophily-speeds}. Conversely, the RFN-AA-I model does not rely on direct inheritance and can circumvent the issue.

\subsubsection{Attention Weights}
A principal way an \gls{rfn} can create sharp boundaries between areas is through our proposed attention mechanism.
We therefore inspect the attention weights w.r.t.\ to the predictions in \cref{tab:volatile-homophily-speeds} to gain insight into the \gls{rfn} prediction process.

The attention weights of the RFN-AA-I model are hard to interpret due to the highly non-linear interactional fusion function.
The fusion function is also capable of selecting information from relations, although at a much finer level.
In general, however, we have noticed a tendency for the model to assign large weights to neighbor segments that differ substantially from the source segment.

\cref{tab:attention-weights-cd} shows the attention weights when predicting segment BD in the Danalien case. 
The representations of adjacent segments are given near-equal weights in the first layer, but in the second layer segment DE and DF have a combined weight of $63.5\%$. This is interesting because the BD segment is part of the residential region, whereas segments DE and DF belong to the transportation region.
We suspect that the model weigh these segments highly because they help identify segment BD as a boundary segment between the residential and transportation regions in the Danalien case.

As shown in \cref{tab:attention-weights-jl}, the model weights segment IJ highly at both layers when predicting the driving speed of segment JL.\
This is interesting because IJ is a motorway segment whereas segment JL is not.
It appears that the model attempts to identify JL as a boundary segment.
However, when predicting the driving speed for its in-neighbor, segment IJ, the attention weights are near-equal at both layers, as shown in \cref{tab:attention-weights-ij}.
This suggests that the identification as a boundary segment is only one-way.
We speculate that the model may sacrifice performance on boundary segments since they are in the minority, for better performance on surrounding segments.

\section{Conclusion}\label{sec:conclusion}
We report on a study of \glspl{gcn} from the perspective of machine learning on road networks.
We argue that many of built-in assumptions of existing proposals do not apply in the road network setting, in particular the assumption of smooth homophily in the network.
In addition, state-of-the-art \glspl{gcn} can leverage only one source of attribute information, whereas we identify three sources of attribute information in road networks: node, edge, and between-edge attributes.
To address these short-comings we propose the \emph{\glsfirst{rfn}}, a novel type of \gls{gcn} for road networks.

We compare different \gls{rfn} variants against a range of baselines, including two state-of-the-art \gls{gcn} algorithms on two machine learning tasks in road networks: driving speed estimation and speed limit classification.
We find that our proposal is able to outperform the \glspl{gcn} baselines by $32\text{--}40\%$ and $21\text{--}24\%$ on driving speed estimation and speed limit classification, respectively. In addition, we report experimental evidence that the assumptions of \glspl{gcn} do not apply to road networks: on the speed limit classification task, the \gls{gcn} baselines failed to outperform a multi-layer perceptron which does not incorporate information from adjacent edges.

In future work, it is of interest to investigate to which extent \glspl{rfn} are capable of transferring knowledge from, e.g., one Danish municipality to the rest of Denmark, given that the inductive nature of our algorithm allows \glspl{rfn} on one road network to enable prediction on another.
If so, it would suggest that \gls{rfn} may be able to learn traffic dynamics that generalize to unseen regions of the network.
This may make it easier to train \glspl{rfn} with less data, but also give more confidence in predictions in, e.g., one of the many regions in the road network of Aalborg Municipality that are labeled sparsely with speed limits. 
In addition, \glspl{rfn} do not incorporate temporal aspects although many road networks tasks are time-dependent. For instance, driving speed estimation where we explicitly excluded driving speeds during peak-hours from our experiments. Extending \glspl{rfn} to learn temporal road network dynamics, e.g., through time-dependent fusion functions that accept temporal inputs, is an important future direction.



\begin{acks}
  The research presented in this paper is supported in part by the DiCyPS project and by grants from the \grantsponsor{obel}{Obel Family Foundation}{} and the \grantsponsor{villum}{Villum Fonden}{}.
We also thank the \glsfirst{osm} contributors who have made this work possible.
Map data is copyrighted by the OpenStreetMap contributors and is available from {https://www.openstreetmap.org}.
\end{acks}

\bibliographystyle{ACM-Reference-Format}
\citestyle{acmnumeric}
\bibliography{references.bib}

\end{document}